# Improved sparse PCA method for face and image recognition

John Jung

CREATORY

john@creatory.vn

Hoa Dieu Nguyen

CREATORY

hoa.nguyen@creatory.vn

Loc Hoang Tran

CREATORY

tran0398@umn.edu

Abstract: Face recognition is the very significant field in pattern recognition area. It has multiple applications in military and finance, to name a few. In this paper, the combination of the sparse PCA with the nearest-neighbor method (and with the kernel ridge regression method) will be proposed and will be applied to solve the face recognition problem. Experimental results illustrate that the accuracy of the combination of the sparse PCA method (using the proximal gradient method and the FISTA method) and one specific classification system may be lower than the accuracy of the combination of the PCA method and one specific classification system but sometimes the combination of the sparse PCA method (using the proximal gradient method or the FISTA method) and one specific classification system leads to better accuracy. Moreover, we recognize that the process computing the sparse PCA algorithm using the FISTA method is always faster than the process computing the sparse PCA algorithm using the proximal gradient method.

I.  Introduction

Face recognition is one type of bio-metric security systems. It has a lot of applications in finance and military. In the years 1990, computer scientists have employed the Eigenface [1] to recognize faces. The Eigenface method utilized the Principle Component Analysis (PCA) which is one of the dimensional reduction methods [2, 3] to reduce the dimensions of the data in order to remove noise and redundant features in the data. This will lead to the low time complexity of the face recognition system. The PCA method has been utilized in a lot of applications such as speech recognition [4, 5]. Then, the Eigenface used the nearest-neighbor method [6] to recognize faces.

To recognize faces, the machine learning approach [7, 8, 9, 10, 11, 12] can also be employed such as the Support Vector Machine (SVM) method. From 2000 to 2010, the Support Vector Machine learning method can be considered the state of the art machine learning method. Please note that the kernel ridge regression method can be viewed as the simplest form of the Support Vector Machine learning method. In this paper, in addition to the nearest-neighbor method, we will also employ the kernel ridge regression method to recognize faces.

Please note that the PCA has two major weaknesses which are the nonexistence of sparsity of the loading vectors and each principle component is the linear combination of all features. To overcome these weaknesses and to present sparsity, many methods have been suggested such as [13, 14, 15, 16]. In this paper, we will introduce two new approaches solving sparse PCA using the Proximal Gradient method [17, 18, 19], and the Fast Iterative Shrinkage-Thresholding method (i.e. FISTA method) [17, 18, 19]. Finally, we will try to combine the sparse PCA dimensional reduction method with the nearest-neighbor method (and with the kernel ridge regression method) and use these combinations to solve the face recognition problem.

We will organize the paper as follows: Section II will present the detailed version of the sparse PCA algorithm using the proximal gradient method in detail. Section III will present the detailed version of the sparse PCA algorithm using the FISTA method in detail. In section IV, we will apply the combination of the sparse PCA algorithm with the nearest-neighbor algorithm and with the kernel ridge regression algorithm to faces in the dataset available from [17]. Section V will conclude this paper and discuss the future directions of researches of this face recognition problem.

II. The proximal gradient method

Suppose that we would like to solve the unconstrained optimization problem like the following

$$min_z f(z)$$

One of the simplest methods solving the above unconstrained optimization problem is the gradient method. This gradient method solve the above optimization as follows

$$z_0 \in R^n, z_{k+1} = z_k - t_{k+1}\nabla f(z_k),$$

where $t_{k+1}$ is the suitable step size.

The idea of this gradient method can also be applied to the $l_1$-regularization problem like the following.

Suppose that we are given a function $f$ that can be decomposed into two functions as follows

$$f(x) = g(x) + \lambda h(x)$$

For this decomposable function $f$, we want to find the solution of the following optimization problem

$$\min_z f(z) = g(z) + \lambda h(z)$$

Please note that function $g$ is differentiable.

So, the solution of this optimization is

$$x^+ = argmin_z g(x) + \nabla g(x)^T(z - x) + \frac{1}{2t}||z - x||_2^2 + \lambda h(z)$$

$$= argmin_z \frac{1}{2t}||z - (x - t\nabla g(x))||_2^2 + \lambda h(z)$$

$$= prox_{h,\lambda t}(x - t\nabla g(x))$$

Please note that $\nabla^2 g(x)$ is replaced by $\frac{1}{t}I$.

This method can also be called the ISTA method. In the next section, we will present the FISTA method which is the extended version of the ISTA method and is a lot faster than the ISTA method.

In our sparse PCA problem, $g(x) = -x^T D^T D x$ and $h(x) = ||x||_1$.

So, we know that $\nabla g(x) = -2D^T Dx$.

Hence the solution of the sparse PCA problem is

$$x^+ = prox_{h,\lambda t}(x + 2tD^T Dx)$$

$$= prox_{h,\lambda t}((I + 2tD^T D)x)$$

Last but not least, in detail, the proximal operator is defined like the following

$$prox_{h,\lambda t}(x - t\nabla g(x))_i = (|x + 2tD^T Dx|_i - \lambda t)_+ sign((x + 2tD^T Dx)_i)$$

We repeat the above formula until it converges to the final solution.

In the next part, we will give the algorithm of one pass of the proximal gradient method (i.e. the ISTA method).

---

Algorithm 1: One pass of proximal gradient method

---

1. Start with $x_k$
2. Compute $x_{k+1} = prox_{h,\lambda t}((I + 2tD^T D)x_k)$
3. $x_{k+1}$ will be served as the input for the next pass of the ISTA method

III. The FISTA method

In this section, we will describe the modified version of the proximal gradient method (i.e. FISTA method) used to solve the sparse PCA problem.

First, randomly choose $t\_old\_old$, $t\_old$, $x\_old\_old$, $x\_old$.

Then, we need to compute

$$y = x_{old} + \frac{t_{old\_old} - 1}{t_{old}}(x_{old} - x_{old\_old})$$

Next, we need to compute the solution of the following formula

$$x^+ = prox_{h,\lambda t}(y - t\nabla g(y))$$

Next, we need to compute $t = \dfrac{1+\sqrt{1+4t_{old}^2}}{2}$

Finally, set *t_old* =*t*, *t_old_old* =*t_old* , *x_old* =$x^+$, *x_old_old* =*x_old* .

We repeat the above procedure until it converges to the final solution.

---

Algorithm 2: One pass of FISTA method

---

1. Start with *t_old_old*, *t_old*, *x_old_old*, *x_old*
2. Compute

$$y = x_{old} + \frac{t_{old\_old} - 1}{t_{old}}(x_{old} - x_{old\_old})$$

3. Compute $x_{k+1} = prox_{h,\lambda t}((I + 2tD^TD)y)$

4. Compute $t = \dfrac{1+\sqrt{1+4t_{old}^2}}{2}$

5. $x_{k+1}$, *x_old*, *t*, *t_old* will be served as the inputs for the next pass of the FISTA method

IV.    Experiments and Results

In this paper, the set of 120 face samples recorded of 15 different people (8 face samples per people) is the training set. Then another set of 45 face samples of these people is the testing set. This dataset is available from [20]. Then, we will merge all rows of the face sample (i.e. the matrix) sequentially from the first row to the last row into a single big row which is the $R^{1*1024}$ row vector. These row vectors will be used as the feature vectors of the nearest-neighbor method and the kernel ridge regression method.

Next, the PCA and the sparse PCA algorithms will be applied to faces in the training set and the testing set to reduce the dimensions of the faces. Then the nearest-neighbor method and the kernel ridge regression method will be applied to these new transformed feature vectors.

In this section, we experiment with the above nearest-neighbor method and kernel ridge regression method in terms of accuracy. The accuracy measure Q is given as follows:

$$Q = \frac{True\ Positive + True\ Negative}{True\ Positive + True\ Negative + False\ Positive + False\ Negative}$$

All experiments were implemented in Matlab 6.5 on virtual machine. The accuracies of the above proposed methods are given in the following table 1 and table 2 and the time complexities of the processes computing the sparse PCA using the proximal gradient and the FISTA methods are given in the following table 3.

Table 1: **Accuracies** of the nearest-neighbor method, the combination of PCA method and the nearest-neighbor method, and the combination of sparse PCA method and the nearest-neighbor method

| | Accuracy (%) |
|---|---|
| The nearest-neighbor method | 90.81 |
| PCA (d = 200) + The nearest-neighbor method | 91.11 |
| PCA (d = 300) + The nearest-neighbor method | 91.11 |
| PCA (d = 400) + The nearest-neighbor method | 91.11 |
| PCA (d = 500) + The nearest-neighbor method | 91.11 |
| PCA (d = 600) + The nearest-neighbor method | 91.11 |
| Proximal Gradient Sparse PCA (d = 200) + The nearest-neighbor method | 87.26 |
| Proximal Gradient Sparse PCA (d = 300) + The nearest-neighbor method | 87.56 |
| Proximal Gradient Sparse PCA (d = 400) + The nearest-neighbor method | 87.85 |

| | Accuracy (%) |
|---|---|
| Proximal Gradient Sparse PCA (d = 500) + The nearest-neighbor method | 87.56 |
| Proximal Gradient Sparse PCA (d = 600) + The nearest-neighbor method | 87.26 |
| FISTA Sparse PCA (d = 200) + The nearest-neighbor method | 88.15 |
| FISTA Sparse PCA (d = 300) + The nearest-neighbor method | 88.44 |
| FISTA Sparse PCA (d = 400) + The nearest-neighbor method | 88.74 |
| FISTA Sparse PCA (d = 500) + The nearest-neighbor method | 89.04 |
| FISTA Sparse PCA (d = 600) + The nearest-neighbor method | 88.74 |

Table 2: **Accuracies** of the kernel ridge regression method, the combination of PCA method and the kernel ridge regression method, and the combination of sparse PCA method and the kernel ridge regression method

| | Accuracy (%) |
|---|---|
| The kernel ridge regression method | 95.85 |
| PCA (d = 200) + The kernel ridge regression method | 96.15 |
| PCA (d = 300) + The kernel ridge regression method | 96.15 |
| PCA (d = 400) + The kernel ridge regression method | 96.15 |
| PCA (d = 500) + The kernel ridge regression method | 96.15 |
| PCA (d = 600) + The kernel ridge regression method | 96.15 |
| Proximal Gradient Sparse PCA (d = 200) + The kernel ridge regression method | 95.26 |
| Proximal Gradient Sparse PCA (d = 300) + The kernel ridge regression method | 96.15 |
| Proximal Gradient Sparse PCA (d = 400) + The kernel ridge regression method | **96.44** |
| Proximal Gradient Sparse PCA (d = 500) + The kernel ridge regression method | 96.15 |
| Proximal Gradient Sparse PCA (d = 600) + The kernel ridge regression method | 96.15 |
| FISTA Sparse PCA (d = 200) + The kernel ridge regression method | 87.26 |
| FISTA Sparse PCA (d = 300) + The kernel ridge regression method | 89.93 |
| FISTA Sparse PCA (d = 400) + The kernel ridge regression method | 90.52 |
| FISTA Sparse PCA (d = 500) + The kernel ridge regression method | 92.30 |
| FISTA Sparse PCA (d = 600) + The kernel ridge regression method | 91.11 |

Table 3: **The time complexities** of the processes computing the PCA and the sparse PCA using the proximal gradient and the FISTA methods

| The time complexities (seconds) | |
|---|---|
| PCA (d = 200) | 8.09 |
| PCA (d = 300) | 8.54 |
| PCA (d = 400) | 9.91 |
| PCA (d = 500) | 12.37 |
| PCA (d = 600) | 14.70 |
| Proximal Gradient Sparse PCA (d = 200) | 2232 |
| Proximal Gradient Sparse PCA (d = 300) | 3117 |
| Proximal Gradient Sparse PCA (d = 400) | 3537 |
| Proximal Gradient Sparse PCA (d = 500) | 4837 |
| Proximal Gradient Sparse PCA (d = 600) | 5370 |
| FISTA Sparse PCA (d = 200) | 439 |
| FISTA Sparse PCA (d = 300) | 806 |
| FISTA Sparse PCA (d = 400) | 1210 |
| FISTA Sparse PCA (d = 500) | 1489 |
| FISTA Sparse PCA (d = 600) | 1634 |

From the above tables 1,2, and 3, we recognize the accuracies of the combination of the sparse PCA method (using the proximal gradient method and the FISTA method) and one specific classification system may be lower than the accuracy of the combination of the PCA method and one specific classification system (may be lower than ~ 4-9%) but sometimes the combination of the sparse PCA method (using the proximal gradient method and the FISTA method) and one specific classification system lead to better accuracy. Because we choose the initial vector of the proximal gradient method (i.e. the ISTA method) and the FISTA method randomly, this will not lead to the best accuracy performance measures of these two methods.

Last but not least, from the experiments, the process computing the sparse PCA algorithm using FISTA method is a lot faster than the process computing the sparse PCA algorithm using proximal gradient method.

V. Conclusions

In this paper, the detailed versions of the sparse PCA method solved by the proximal gradient method and the FISTA method have been proposed. The experimental results show that the accuracy of the combination of the sparse PCA method (using the proximal gradient method and the FISTA method) and one specific classification system may be lower than the accuracy of the combination of the PCA method and one specific classification system but sometimes the combination of the sparse PCA method (using the proximal gradient method and the FISTA method) and one specific classification system leads to better accuracy.

In the future, we will test the accuracies of the combination of the tensor sparse PCA method with a lot of classification systems, for e.g. the SVM method and deep neural network method, for not only the face recognition problem but also other image recognition problems such as finger print recognition problem.

## References

1. Turk, Matthew, and Alex Pentland. "Eigenfaces for recognition." *Journal of cognitive neuroscience* 3.1 (1991): 71-86.
2. Tran, Loc Hoang, Linh Hoang Tran, and Hoang Trang. "Combinatorial and Random Walk Hypergraph Laplacian Eigenmaps." *International Journal of Machine Learning and Computing* 5.6 (2015): 462.
3. Tran, Loc, et al. "WEIGHTED UN-NORMALIZED HYPERGRAPH LAPLACIAN EIGENMAPS FOR CLASSIFICATION PROBLEMS." *International Journal of Advances in Soft Computing & Its Applications* 10.3 (2018).
4. Trang, Hoang, Tran Hoang Loc, and Huynh Bui Hoang Nam. "Proposed combination of PCA and MFCC feature extraction in speech recognition system." *2014 International Conference on Advanced Technologies for Communications (ATC 2014)*. IEEE, 2014.
5. Tran, Loc Hoang, and Linh Hoang Tran. "The combination of Sparse Principle Component Analysis and Kernel Ridge Regression methods applied to speech recognition problem." *International Journal of Advances in Soft Computing & Its Applications* 10.2 (2018).
6. Zhang, Zhongheng. "Introduction to machine learning: k-nearest neighbors." *Annals of translational medicine* 4.11 (2016).
7. Scholkopf, Bernhard, and Alexander J. Smola. *Learning with kernels: support vector machines, regularization, optimization, and beyond*. MIT press, 2001.
8. Trang, Hoang, and Loc Tran. "Kernel ridge regression method applied to speech recognition problem: A novel approach." *2014 International Conference on Advanced Technologies for Communications (ATC 2014)*. IEEE, 2014.

9. Zurada, Jacek M. *Introduction to artificial neural systems*. Vol. 8. St. Paul: West publishing company, 1992.
10. Tran, Loc. "Application of three graph Laplacian based semi-supervised learning methods to protein function prediction problem." *arXiv preprint arXiv:1211.4289* (2012).
11. Tran, Loc, and Linh Tran. "The Un-normalized Graph p-Laplacian based Semi-supervised Learning Method and Speech Recognition Problem." *International Journal of Advances in Soft Computing & Its Applications* 9.1 (2017).
12. Trang, Hoang, and Loc Hoang Tran. "Graph Based Semi-supervised Learning Methods Applied to Speech Recognition Problem." *International Conference on Nature of Computation and Communication*. Springer, Cham, 2014.
13. R.E. Hausman. Constrained multivariate analysis. Studies in the Management Sciences, 19 (1982), pp. 137–151
14. Vines, S. K. "Simple principal components." *Journal of the Royal Statistical Society: Series C (Applied Statistics)* 49.4 (2000): 441-451.
15. Jolliffe, Ian T., Nickolay T. Trendafilov, and Mudassir Uddin. "A modified principal component technique based on the LASSO." *Journal of computational and Graphical Statistics* 12.3 (2003): 531-547.
16. Zou, Hui, Trevor Hastie, and Robert Tibshirani. "Sparse principal component analysis." *Journal of computational and graphical statistics* 15.2 (2006): 265-286.
17. Tao, Shaozhe, Daniel Boley, and Shuzhong Zhang. "Local linear convergence of ISTA and FISTA on the LASSO problem." *SIAM Journal on Optimization* 26.1 (2016): 313-336.
18. Beck, Amir, and Marc Teboulle. "A fast iterative shrinkage-thresholding algorithm for linear inverse problems." *SIAM journal on imaging sciences* 2.1 (2009): 183-202.
19. Daubechies, Ingrid, Michel Defrise, and Christine De Mol. "An iterative thresholding algorithm for linear inverse problems with a sparsity constraint." *Communications on Pure and Applied Mathematics: A Journal Issued by the Courant Institute of Mathematical Sciences* 57.11 (2004): 1413-1457.
20. http://www.cad.zju.edu.cn/home/dengcai/Data/FaceData.html